\documentclass[sigconf]{acmart}
\AtBeginDocument{%
  }

\setcopyright{acmlicensed}
\copyrightyear{2026}
\acmYear{2026}
\setcopyright{cc}
\setcctype{by}
\acmConference[KDD 2026] {Proceedings of the 32nd ACM SIGKDD Conference on Knowledge Discovery and Data Mining V.2}{August 9--13, 2026}{Jeju Island, Republic of Korea.}
\acmBooktitle{Proceedings of the 32nd ACM SIGKDD Conference on Knowledge Discovery and Data Mining V.2 (KDD 2026), August 9--13, 2026, Jeju Island, Republic of Korea}

\acmDOI{10.1145/3770855.3817465}
\acmConference[KDD 2026] {Proceedings of the 32nd ACM SIGKDD Conference on Knowledge Discovery and Data Mining V.2}{August 9--13, 2026}{Jeju Island, Republic of Korea.}
\acmISBN{979-8-4007-2259-2/2026/08}

\usepackage{caption}
\usepackage{multirow}
\usepackage{subcaption}
\usepackage{natbib}
\usepackage{hyperref}
\usepackage[capitalize]{cleveref}
\usepackage{stfloats}
\usepackage{xcolor} 
\usepackage{balance}
\definecolor{LightPink}{HTML}{FFB6C1}    
\definecolor{DeepPink}{HTML}{FF1493}     
\definecolor{SakuraPink}{HTML}{FEE2E2}   
\crefname{section}{appendix}{appendices}

\begin{document}
\title{Evaluating Uplift Modeling under Structural Biases: Insights into Metric Stability and Model Robustness}

\author{Yuxuan Yang}
\affiliation{%
\department{College of Computer Science and Software Engineering}
  \institution{Shenzhen University}
  \city{Shenzhen}
  \state{Guangdong}
  \country{China}
}
\email{yyuxuan959@gmail.com}

\author{Dugang Liu}
\authornote{Co-corresponding Authors.} 
\affiliation{%
\department{College of Computer Science and Software Engineering}
  \institution{Shenzhen University}
  \city{Shenzhen}
  \state{Guangdong}
  \country{China}
}
\email{dugang.ldg@gmail.com}

\author{Yiyan Huang}
\authornotemark[1] 
\affiliation{%
\department{School of Computing and Information Technology}
 \institution{Great Bay University}
 \city{Dongguan}
 \state{Guangdong}
 \country{China}}
\email{huangyiyan@gbu.edu.cn}

\renewcommand{\shortauthors}{Yuxuan Yang, Dugang Liu, and Yiyan Huang}

\begin{abstract}
 	In personalized marketing, uplift models estimate the incremental effect of an intervention by modeling how customer behavior would change under alternative treatments using counterfactual analysis. However, real-world marketing data often exhibit various biases, such as selection bias, spillover effects, measurement error, and unobserved confounding. These biases can adversely affect both the accuracy of uplift estimation and the validity of evaluation metrics. Despite the importance of bias-aware assessment, there remains a lack of systematic studies evaluating how different models and metrics perform under such biased conditions. To bridge this gap, we design a systematic benchmarking framework. Unlike standard predictive tasks, real-world uplift datasets inherently lack counterfactual ground truth. This limitation renders the direct validation of evaluation metrics infeasible and prevents the precise quantification of biases. Therefore, a semi-synthetic approach serves as a critical enabler for systematic benchmarking. This approach effectively bridges the gap by retaining real-world feature dependencies while providing the ground truth needed to isolate structural biases. Our investigations reveal that (i) uplift targeting and prediction can manifest as distinct objectives, where proficiency in one does not ensure efficacy in the other; (ii) while many models exhibit inconsistent performance under diverse biases, TARNet shows notable robustness, providing insights for subsequent model design; (iii) the stability of evaluation metrics is linked to their mathematical alignment with the ATE, suggesting that ATE-approximating metrics yield more consistent model rankings under structural data imperfections. These findings suggest the need for more robust uplift models and evaluation metrics under real-world data imperfections.

\end{abstract}





\maketitle

\section{Introduction}
Uplift modeling has emerged as a pivotal methodology for optimizing decision-making in personalized marketing by estimating the individual uplift from target interventions (e.g., advertisements or promotions) for those who are most likely to show incremental responses (e.g., clicks or purchases). Based on the potential outcomes framework of causal inference \citep{rubin1974estimating}, uplift modeling aims to quantify the difference between an individual's potential outcomes, $Y^1$ (under treatment) and $Y^0$ (under control), thereby capturing the individualized treatment effect (ITE) $Y^1 - Y^0$. However, this is challenging due to the fundamental problem in causal inference that only one potential outcome is observable (factual), while the other remains unobservable (counterfactual), rendering the ITE unidentifiable and complicating both model training and evaluation \citep{rubin1974estimating}. 

To address the unidentifiability of ITE, most uplift methods focus on estimating the Conditional Average Treatment Effect (CATE), $\tau(x) = \mathbb{E}[Y^1 - Y^0 \mid X = x]$, which approximates the ITE given observed covariates $X$. State-of-the-art approaches range from meta-learners (e.g., \cite{nie2021quasi,kennedy2023towards}) to neural network architectures (e.g., \cite{shalit2017estimating,shi2019adapting}). This paradigm has been widely employed across diverse domains, including healthcare \citep{jaskowski2012uplift,nassif2013score}, education \citep{olaya2020uplift,tanai2025customize}, policy evaluation \citep{bermeo2023estimation,barile2024causal}, customer retention \citep{devriendt2021you,verhelst2023churn}, recommendation systems \citep{sato2019uplift,wang2024uplift}, and e-commerce \citep{huang2021causal,albert2022commerce,huang2024entire}. Uplift models are particularly critical in data-driven decision-making contexts such as digital marketing, owing to their ability to optimize resource allocation by quantifying incremental effects and maximizing response rates.

Nevertheless, the effectiveness of uplift models depends on several key assumptions in causal inference. First, the Stable Unit Treatment Value Assumption (SUTVA) precludes interference between units. Second, the unconfoundedness assumption requires that all covariates affecting both treatments and outcomes be fully observed. Third, uplift models are typically trained on real-world marketing data under randomized controlled trials (RCTs), thereby mitigating covariate shift between treatment and control groups. In practice, these assumptions often fail due to common biases in marketing data—such as selection bias, spillover effects, measurement error, and hidden confounding—which can distort uplift estimates and undermine decision-making \citep{luo2024survey,ali2021measuring,chen2024conformal}.

Similarly, uplift model evaluation faces parallel challenges. The gold-standard metrics, Precision in Estimation of Heterogeneous Effects (PEHE) for individual-level uplift accuracy and Average Treatment Effect (ATE) for population-level targeting performance, are infeasible in practice due to unobservable counterfactuals \citep{wang2023optimal,huang2024unveiling}. Consequently, practitioners rely on surrogate metrics, including Uplift score, Area Under the Uplift Curve (AUUC), and Qini coefficient, to assess model quality \citep{devriendt2018literature,liu2024benchmarking}. These metrics, originally designed for idealized experimental settings, may implicitly depend on the above assumptions. This raises concerns about their reliability in model evaluation when biases violate these constraints, potentially leading to erroneous model rankings and suboptimal model selection \citep{zhurethinking,mahajan2024empirical}.

\textbf{Contributions.} Despite the prevalence of structural biases in online marketing data, their impact on model reliability and metric validity in uplift modeling has received little attention. In this paper, we prioritize establishing a systematic understanding of uplift model and metric robustness under real-world marketing biases, rather than introducing novel methodologies. Our goal is to deliver actionable insights into existing tools, addressing a critical need in uplift modeling research to facilitate reliable deployment in practice and inspire future work on bias-resilient evaluation metrics. Specifically, our contributions are threefold:

\noindent \textbf{(1)} We conduct a comprehensive analysis of the complex interplay between data biases and uplift modeling. We discuss inherent challenges within marketing contexts posed by selection bias, spillover effects, measurement error, and hidden confounding, and provide empirical evidence on their separate impacts on uplift model training and evaluation.

\noindent \textbf{(2)} We propose desiderata for experimental design that disentangle bias effects in uplift benchmarks. We advocate for controlled semi-synthetic environments that simulate realistic marketing scenarios, incorporate tunable bias parameters for isolated analysis, and explore interactions between biases, targeting fractions, uplift learners, and evaluation metrics. Our findings highlight the need for a more careful task-dependent experimental design with adequate targeting fractions and evaluation metrics to select targeting strategies in biased scenarios.

\noindent \textbf{(3)} We gain important insights into the robustness of uplift models and evaluation metrics through our comprehensive empirical investigations. Specifically, we conclude that (i) our results demonstrate a divergence between uplift targeting and prediction, where proficiency in one task does not guarantee success in the other; (ii) while many models exhibit inconsistent performance across diverse biases, TARNet shows notable robustness, suggesting that its structural separation of potential outcome modeling provides valuable insights for designing resilient architectures; (iii) the stability of evaluation metrics is linked to their mathematical alignment with the population-level causal operator, indicating that ATE-approximating metrics yield more consistent model rankings under structural data imperfections.

\section{Related Work}
\paragraph{Common Biases in Personalized Marketing.}
Uplift modeling in personalized marketing is susceptible to several structural biases that can violate key causal assumptions and compromise the accuracy of treatment effect estimation. One major source is \textbf{selection bias}, which arises when treatment assignment is systematically influenced by observed covariates (non-RCT scenario). This can lead to covariate distributional shifts between treated and control groups, reducing the generalizability of outcome models across groups and distorting uplift estimates \citep{mcinerney2020counterfactual,luo2024survey,chen2025data}. Another pervasive phenomenon is the \textbf{spillover effect}, also known as interference, which violates the SUTVA by allowing one individual's treatment assignment to affect the outcomes of others. Such violations are common in social platforms like TikTok, where user interactions can propagate treatment effects across social networks \citep{devriendt2018literature,caljon2024optimizing,gubela2024multiple}. A third source of bias is \textbf{measurement error} in covariates, arising from noisy data sources such as self-reported surveys or misrecorded data. These inaccuracies introduce noise into model inputs, which can cause models to overfit to noisy patterns and misspecified causal effects \citep{bagozzi1998representation,ali2021measuring}. Finally, \textbf{unobserved confounding} poses a fundamental threat to identifiability in causal inference. When latent variables (e.g., user mood or private income) influence both treatment assignment and outcomes but are unmeasured, causal estimators can be non-identifiable, thus leading to biased causal estimates from observational data \citep{chen2024conformal,vrtana2023power,zhu2024mitigating}.

\paragraph{Uplift Modeling and Uplift Evaluation.}
Recent research on \textbf{uplift modeling} has focused on two primary classes of methods: meta-learners and neural network–based models. Meta-learners, including the S-learner, T-learner, X-learner, R-learner, U-learner, and DR-learner \citep{kunzel2019metalearners,nie2021quasi,foster2023orthogonal}, estimate the CATE by fitting separate or joint outcome models with tailored loss functions. While these approaches are widely applicable, their performance depends heavily on the estimation quality of the base learners and the validity of key causal assumptions. Neural network models, such as BNN \citep{johansson2016learning}, TARNet and CFRNet \citep{shalit2017estimating}, Dragonnet \citep{shi2019adapting}, FlexTENet \citep{curth2021inductive}, DESCN \citep{zhong2022descn}, and EFIN \citep{liu2023explicit}, leverage deep representation learning to capture complex and nonlinear potential outcome functions, often incorporating balancing objectives to reduce confounding bias. Despite these methodological innovations, \textbf{uplift evaluation} remains a persistent challenge, especially in observational settings where counterfactual outcomes are unavailable. Ideal evaluation criteria such as the PEHE and ATE provide gold-standard accuracy measures of uplift models, but require access to both potential outcomes, which is infeasible in real-world applications \citep{wang2023optimal,huang2024unveiling}. Consequently, practitioners often rely on gain-curve–based metrics, including Uplift score, AUUC, and Qini coefficient \citep{devriendt2018literature,liu2024benchmarking}. Although widely used, these metrics implicitly rely on the same causal assumptions as the models themselves, which might render them invalid when those assumptions are violated.

\section{Problem Setup}
This paper follows the standard potential outcomes framework for uplift modeling. Let $\{(X_i,T_i,Y_i)\}_{i=1}^{N}$ be a dataset of $N$ units, where $X \in \mathbb{R}^{d}$ denotes customer covariates, $T \in \{0, 1\}$ is a binary treatment, and $Y \in \mathbb{R}$ is a continuous outcome. Each unit has two potential outcomes, $Y^1$ and $Y^0$, while only one of them can be observed as the factual outcome $Y = TY^1 + (1 - T)Y^0$. The CATE is defined as $\tau(x) = \mathbb{E}[Y(1) - Y(0) \mid X = x]$, and the goal of uplift modeling is to learn an estimator $\hat{\tau}(x)$ of $\tau(x)$ from observational data, for either uplift prediction (predicting individual-level uplift values) or uplift targeting (targeting top-ranked individuals to maximize the total response). 

Uplift modeling usually relies on several key assumptions:
\begin{enumerate}
\item \textbf{Randomized Controlled Trials (RCTs):} Treatment is randomly assigned in the training and evaluation datasets.
    \item \textbf{Stable Unit Treatment Value Assumption (SUTVA):} A unit’s outcome depends only on the treatment it receives and is unaffected by the treatments assigned to other units.
    \item \textbf{Unconfoundedness:} There are no unobserved confounders, i.e., $(Y^0, Y^1) \perp T \mid X$.
    \item \textbf{Overlap:} Every unit has a non-zero probability of receiving either treatment, i.e., $P(T = 1 \mid X = x) > 0$.
    \item \textbf{Consistency:} The observed outcome equals the potential outcome under the received treatment.
\end{enumerate}
In most studies, Assumptions (4) and (5) hold by design. However, Assumption (1) may be violated by selection bias or the limited resources required to conduct RCTs; Assumption (2) may be violated by interference effects; and Assumption (3) may be violated by measurement error or hidden confounding. Such violations can bias uplift estimation and lead to misleading model evaluations. \textit{\textbf{Therefore, the goal of this study is to investigate the model reliability and metric validity when Assumptions (1-3) are not satisfied.}}

\subsection{Uplift Models}

This section outlines the construction of common uplift learners using observed samples $\{(X_{i}, T_{i}, Y_{i})\}_{i=1}^{N}$. Let $N^T$ and $N^C$ denote the sample sizes in the treatment and control groups, respectively, such that $N = N^T + N^C$. The construction details are as follows.
\begin{itemize}
	\item \textbf{S-learner}: Fit a single model $\hat{\mu}(X,T)$ with predictors $(X,T)$ and response $Y$, then compute
	\begin{equation*}
		\begin{aligned}
				\hat{\tau}_S(X) = \hat{\mu}(X,1) - \hat{\mu}(X,0).
		\end{aligned}
	\end{equation*}
	\item \textbf{T-learner}: Fit separate models $\hat{\mu}_1(X)$ and $\hat{\mu}_0(X)$ on treated $(X^T, Y^T)$ and control $(X^C, Y^C)$ data, respectively:
	\begin{equation*}
		\begin{aligned}
	\hat{\tau}_T(X) = \hat{\mu}_1(X) - \hat{\mu}_0(X).
		\end{aligned}
	\end{equation*}
	\item \textbf{X-learner} \citep{kunzel2019metalearners}: Fit $\hat{\mu}_1(X)$ and $\hat{\mu}_0(X)$ (T-learner), estimate the propensity score $P(T=1 \mid X)$ with $\hat{\pi}(X)$, regress treatment effects for treated and control units, and fit models $\hat{\tau}^1_X$ and $\hat{\tau}^0_X$ separately, then combine:
		\begin{equation*}
		\begin{aligned}
		&\hat{\tau}_{X}(X) = (1 - \hat{\pi}(X))\hat{\tau}^1_{X}(X) + \hat{\pi}(X)\hat{\tau}^0_{X}(X),\\
&\hat{\tau}^1_{X}=\mathop{\arg\min}_{\tau} \; \frac{1}{N^T}\sum_{i=1}^{N^T}(\tau(X_i) - (Y_i -  \hat{\mu}_0(X_i)))^2,\\
&\hat{\tau}^0_{X}=\mathop{\arg\min}_{\tau} \; \frac{1}{N^C}\sum_{i=1}^{N^C}(\tau(X_i) - (\hat{\mu}_1(X_i) - Y_i))^2.
		\end{aligned}
	\end{equation*}
	\item \textbf{R-learner} \citep{nie2021quasi}: Fit a factual outcome model $\hat{\mu}(X)$ and the propensity score model $\hat{\pi}(X)$, and compute residuals $\xi = Y - \hat{\mu}(X)$ and $\nu = T - \hat{\pi}(X)$, then fit $\hat{\tau}_R(X)$ by
			\begin{equation*}
		\begin{aligned}
				\hat{\tau}_{R}=\mathop{\arg\min}_{\tau} \; \frac{1}{N}\sum_{i=1}^{N}(\xi_i - \nu_i\tau(X_i))^2.
		\end{aligned}
	\end{equation*}
	\item \textbf{U-learner} \citep{nie2021quasi}: Fit $\hat{\mu}(X)$ and $\hat{\pi}(X)$, compute residuals $\xi = Y - \hat{\mu}(X)$ and $\nu = T - \hat{\pi}(X)$, then regress $\xi / \nu$ on $X$:
				\begin{equation*}
		\begin{aligned}
			\hat{\tau}_{U}=\mathop{\arg\min}_{\tau} \; \frac{1}{N}\sum_{i=1}^{N}(\frac{\xi_i}{\nu_i}- \tau(X_i))^2.
		\end{aligned}
	\end{equation*}
	
	\item \textbf{DR-learner} \citep{kennedy2023towards, foster2023orthogonal}: Fit $\hat{\mu}_1(X)$, $\hat{\mu}_0(X)$, and $\hat{\pi}(X)$. Construct doubly robust pseudo-outcomes $Y_{DR}^1 = \hat{\mu}_1(X) + \frac{T}{\hat{\pi}(X)}\left(Y - \hat{\mu}_1(X)\right)$ and $Y_{DR}^0 = \hat{\mu}_0(X) + \frac{1-T}{1-\hat{\pi}(X)}\left(Y - \hat{\mu}_0(X)\right)$, then regress $Y_{DR}^1 - Y_{DR}^0$ on $X$:
			\begin{equation*}
		\begin{aligned}
	\hat{\tau}_{DR}=\mathop{\arg\min}_{\tau} \; \frac{1}{N}\sum_{i=1}^{N}(\tau(X_i) - (Y^1_{i,DR} - Y^0_{i,DR}))^2.
		\end{aligned}
	\end{equation*}

	\item \textbf{RA-learner} \citep{curth2021nonparametric}: Fit $\hat{\mu}_1(X)$ and $\hat{\mu}_0(X)$, and construct regression-adjusted pseudo-outcomes $Y_{RA} = T\left(Y - \hat{\mu}_0(X)\right) + (1-T)\left(\hat{\mu}_1(X) - Y\right)$, then regress $Y_{RA}$ on $X$:
				\begin{equation*}
		\begin{aligned}
		\hat{\tau}_{RA} = \mathop{\arg\min}_{\tau} \; \frac{1}{N}\sum_{i=1}^{N}(\tau(X_i) - Y_{i,RA})^2.
		\end{aligned}
	\end{equation*}
	
	\item \textbf{TARNet} \citep{shalit2017estimating}: Learn a shared representation $\Phi(X)$ via a neural network, then use two separate heads $h_1(\Phi)$ and $h_0(\Phi)$ for treated and control outcomes. The CATE is $\hat{\tau}(X) = h_1(\Phi(X)) - h_0(\Phi(X))$.
	
	\item \textbf{Dragonnet} \citep{shi2019adapting}: Builds on TARNet by adding a third head, $h_{\pi}(\Phi(X))$, to predict the propensity score alongside the two outcome heads. Together with the targeted regularization technique, it enables joint optimization of representations, outcome models, and propensity scores to improve average treatment effect estimation.
\end{itemize}

\subsection{Evaluation Metrics} \label{sec:evaluation metric}
We evaluate model performance using five metrics, categorized into two groups: (i) oracle metrics, including PEHE$_k$ \cite{hill2011bayesian,shalit2017estimating,zhurethinking} and ATE$_k$; and (ii) practical metrics, comprising Uplift$_k$, AUUC$_k$ \citep{rzepakowski2012decision,zhurethinking} and Qini$_k$ \citep{surry2011quality,radcliffe2007using,diemert2018large,devriendt2018literature,belbahri2021qini}. Detailed descriptions are provided in \Cref{Appendix B}.

\section{Challenges for Uplift Modeling}
In this section, we discuss three main challenges that online marketing practitioners encounter in uplift modeling.

\noindent \textbf{Challenge 1: Missing Counterfactual Outcomes.}

The core challenge in causal inference is the missing counterfactual. For any individual, we can only observe one potential outcome (either under treatment or under control), but never both \cite{holland1986statistics}. This fundamental limitation precludes direct learning of individual-level treatment effects, complicating both training and evaluation tasks in uplift modeling.

\noindent \textbf{Challenge 2: Structural Biases in Marketing Data.}

\textit{Selection bias: personalized targeting induces covariate shift between treatment groups.}
Personalized marketing frequently leverages recommender systems informed by customer preferences (e.g., browsing behavior \citep{li2018learning} and past purchases \citep{chen2025data}). When recommendations drive treatment assignment, the treated and control groups differ systematically in their covariates. Formally, assignment is non-random and exhibits covariate shift: $P(X \mid T = 1) \neq P(X \mid T = 0)$, which further leads to a shift between factual and counterfactual distributions \citep{shalit2017estimating}. As a result, models trained on the factual domain cannot generalize to the entire domain \citep{tong2025disc2o,curth2023search,jeong2020robust}.

\textit{Spillover effects: social influence violates SUTVA and overestimates uplift values.}
Consumers' purchasing decisions are often affected by factors related to other people (e.g., perceived price fairness \citep{gubela2024multiple}, friends' recommendations \cite{xu2022travelers}), which violates the fundamental SUTVA assumption. In this case, potential outcomes are no longer independent across individuals. For a specific individual $i$, $Y_i$ can depend on treatments and outcomes of their neighbors. This network dependence complicates the potential-outcome structure, making individual covariates insufficient to fully explain uplift values and introducing bias into CATE estimation \citep{hudgens2008toward,pouget2019variance}.

\textit{Measurement error: noisy customer features distort training and estimation.}
Platform-collected features can contain errors due to limited time in research interviews \citep{eliashberg1985measurement}, data entry mistakes \citep{tsikriktsis2005review}, and survey design \citep{ali2021measuring}. Denoting the true features by $X$ and the measurement noise by $\epsilon_{x}$, the observed features become $X_{\mathrm{obs}} = X + \epsilon_x$. Such perturbations degrade feature informativeness and propagate bias into CATE estimates \citep{schennach2016recent,saeed2020anchored}.

\textit{Unobserved confounding: latent factors confound treatment–outcome relationships.}
Unmeasured factors in marketing, such as cultural context \citep{ogden2004exploring} and transient mood \cite{asshidin2016perceived}, can violate the Unconfoundedness assumption, leading to $(Y^0, Y^1) \not\perp T \mid X$. Even with rich observed covariates, missing latent factors still prevent correct model specification, undermining uplift estimation and the validity of downstream decision-making \citep{veitch2019using,louizos2017causal,tchetgen2014control}.

\noindent \textbf{Challenge 3: Metric Validity under Biases.}

PEHE and ATE are ideal metrics for evaluating uplift models \cite{wang2023optimal,huang2024unveiling}, but they are infeasible in practice due to the missing counterfactual problem (Challenge 1). Practitioners therefore rely on observable surrogates such as Uplift, AUUC, and the Qini coefficient (see Section \ref{sec:evaluation metric}), which are typically informative in randomized controlled trials \citep{bokelmann2024improving}. In real marketing settings, however, data are rarely randomized, and even when they are randomized, they often exhibit the biases described above. This raises a critical yet underexplored question: to what extent do selection bias, spillover effects, measurement error, and unobserved confounding compromise these evaluation metrics? Prior work has shown that these metrics can be unreliable in certain cases \cite{renaudin2021evaluation,zhurethinking}, but the field still lacks a clear mapping from the type and magnitude of bias to the resulting degradation in metric reliability. Such an exploration is essential for model selection and for designing robust evaluation protocols in real-world biased environments.

\section{Experiments}
\subsection{Experimental Setup}
To evaluate the robustness of uplift models and their evaluation metrics under structural biases commonly observed in marketing data, we design a semi-synthetic experimental framework, a widely adopted approach in causal inference \citep{curth2023search,mahajan2024empirical,huang2024unveiling}. This is motivated by three considerations. First, real-world datasets lack counterfactual outcomes, making it impossible to directly assess model accuracy or metric validity. Second, although many benchmark uplift datasets originate from RCTs, they may still be affected by spillover effects or measurement error, hindering both uplift estimation and model evaluation. Third, semi-synthetic data generation allows us to control the severity of multiple bias types while preserving realistic feature distributions, enabling comprehensive and systematic robustness analysis.
To create the semi-synthetic data, we collect covariates from the Hillstrom dataset \cite{hillstrom2008minethatdata}, a benchmark in uplift modeling \citep{soltys2015ensemble,zhang2021unified}. Compared to the large-scale and sparse Criteo dataset \cite{Diemert2018}, Hillstrom’s compact scale allows for better experimental control. By mitigating the interference of high-dimensional sparsity, this dataset provides an effective testbed for a focused analysis of our proposed mechanisms and the resulting performance shifts. Specifically, it contains $n=64{,}000$ samples, and each sample has covariates $X_i \in \mathbb{R}^d$ with $d=8$, consisting of 1 continuous and 7 discrete variables. We simulate treatment assignments and potential outcomes using the following data-generating process (DGP). All experimental details and additional results are available via \url{https://github.com/Uplift-Bench/Uplift_Evaluation_Bench}.

We first define the neighborhood of unit $i$ as
\[
N(i) = \{X_j : \|X_i - X_j\|_2 \leq 0.1\},
\]
which captures all units within a fixed Euclidean distance. Using $N(i)$, we compute the average treatment value and the average outcome value contributed by its neighbors:
\begin{equation*}
	\sigma_{N(i)} = \frac{1}{|N(i)|} \sum_{X_j \in N(i)} \zeta_j, \quad
	\gamma_{N(i)}^t = \frac{1}{|N(i)|} \sum_{X_j \in N(i)} \gamma_j^t.
\end{equation*}
The treatment assignment for unit $i$ follows a Bernoulli distribution:
\begin{equation}
	T_i \mid X_i \sim \mathrm{Bern} \left( 1/\left(1 + \exp\left(-\xi \left(\zeta_i + 0.2\, \sigma_{N(i)} + 0.3\right)\right)\right) \right),
\end{equation}
where the individual baseline treatment value function is defined as $\zeta_i = \beta_T^\top X_i$ with $\beta_T \sim \mathcal{N}(-0.2, 0.01)$.

The potential outcomes and observed outcomes are generated by
\begin{equation}
	\begin{aligned}
		&Y_i^t = \gamma_{i}^t + \theta_t \gamma_{N(i)}^t, && t \in \{0,1\};\\
		&Y_i = T_iY_i^1 + (1 - T_i)Y_i^0  + \epsilon, &&\epsilon \sim \mathcal{N}(0, 0.1),
	\end{aligned}
\end{equation}
where $\gamma_{i}^t$ captures the $i$-th individual’s baseline outcome value under treatment $t$, and $\gamma_{N(i)}^t$ represents the average $\gamma^t$ of its neighbors. Specifically, the function $\gamma_{i}^t$ incorporates linear, quadratic, and cubic (for $t = 1$) interaction terms of the covariates:
\begin{equation*}
	\begin{aligned}
		&\gamma_{i}^0 = \sum_{j=1}^d \beta_j^0 X_{i,j} 
		+ \sum_{j,k=1}^d \beta_{j,k}^0 X_{i,j} X_{i,k},\\
		&\gamma_{i}^1 = \sum_{j=1}^d \beta_j^1 X_{i,j} 
		+ \sum_{j,k=1}^d \beta_{j,k}^1 X_{i,j} X_{i,k}
		+ \sum_{j,k,l=1}^d \beta_{j,k,l}^1 X_{i,j} X_{i,k} X_{i,l}.
	\end{aligned}
\end{equation*}
The coefficients are drawn as follows: $\beta_j^0 \sim \mathrm{Bern}(0.3)$, $\beta_{j,k}^0 \sim \mathrm{Bern}(0.2)$, $\beta_j^1 \sim \mathrm{Bern}(0.2)$, $\beta_{j,k}^1 \sim \mathrm{Bern}(0.5)$, and $\beta_{j,k,l}^1 \sim \mathrm{Bern}(0.6)$. Moreover, this DGP allows us to systematically control four types of structural biases through the following dedicated parameters:
\begin{itemize}
\item Selection Bias: controlled via $\xi$, with values $\{0.8, 1.6, 2.4\}$. Other parameters are fixed at $m = 0.1$, $\omega = 1.2$, and $(\theta_0, \theta_1) = (0.4, 0.8)$.
\item Spillover Effects: induced through $(\theta_0, \theta_1)$, which determine the strength of network effects in the outcome models, taking values $\{(0.4, 0.8), (0.5, 0.95), (0.6, 1.1)\}$. Remaining parameters are fixed at $\xi = 0$, $m = 0.1$, and $\omega = 1.2$.
\item Measurement Error: introduced by adding noise to the covariates, which introduces perturbations into observational covariates $X_{\mathrm{obs}} = X + \epsilon_x$ with $\epsilon_x \sim \mathcal{N}(0, \omega/8)$, where $\omega \in \{1.2, 2.4, 3.6\}$. Other parameters are fixed at $\xi = 0$, $m = 0.1$, and $(\theta_0, \theta_1) = (0.4, 0.8)$.
\item Unobserved Confounding: simulated by removing a fraction $m$ of covariates, reducing the observed feature dimension to $d - \lfloor m d \rfloor$. The knob $m$ takes values $\{0.1, 0.3, 0.5\}$, representing levels of hidden confounding. Other parameters are fixed at $\xi = 0$, $\omega = 1.2$, and $(\theta_0, \theta_1) = (0.4, 0.8)$.
\end{itemize}
Note that larger knob values correspond to stronger biases in all four bias settings. For each setting, all results are averaged over 10 independent runs on the fixed dataset to mitigate the impact of random initializations. The dataset is split into training, validation, and test sets with a ratio of $49\%/21\%/30\%$.

\subsection{Model Training and Evaluation}

\;\;\;\textit{Model Training.} We train 9 uplift modeling approaches, covering both meta-learning algorithms (S-, T-, X-, R-, U-, DR-, RA-learners) and neural network architectures (TARNet and Dragonnet). All meta-learners share the same model backbones: we use XGBoost (eXtreme Gradient Boosting) for outcome estimation and LR (logistic regression) for propensity score estimation. All training models are tuned using a combination of Optuna \citep{akiba2019optuna} and grid search to ensure fair comparison. The details of the hyperparameter search space for LR, XGBoost, TARNet, and Dragonnet are provided in \Cref{hyperparameter}. All model training processes are conducted on a Dell 3640 workstation with an Intel Xeon W-1290P 3.60 GHz CPU and an NVIDIA GeForce RTX 3080 Ti GPU.

\textit{Model Evaluation.} To evaluate model performance, we adopt 5 widely used uplift evaluation metrics that reflect different aspects of model performance. For oracle evaluation metrics (counterfactual outcomes are available), we use PEHE$_k$ to measure individual-level estimation accuracy (lower is better), and ATE$_k$ to measure population-level targeting profit (higher is better). For utility-oriented evaluation (only observed data are available), we consider Uplift$_k$, AUUC$_k$, and Qini$_k$, with the values of AUUC$_k$ and Qini$_k$ scaled by $N_k$ (higher is better for all three metrics).

The following sections present a comprehensive empirical analysis, focusing on two main aspects: 
(i) \textbf{model performance under different structural biases}; and 
(ii) \textbf{robustness of common evaluation metrics under assumption violations}. 
All results are reported on the test sets.

\begin{table}[H]
	\centering
	\caption{The search space of hyperparameters. LR and XGBoost follow the scikit-learn implementations. TARNet and Dragonnet follow the implementations in \cite{shalit2017estimating, shi2019adapting}.}
        \label{hyperparameter}
		\resizebox{1\columnwidth}{!}{
    \begin{tabular}{cccc}
	&       &       &  \\
	\midrule
	Model & Parameter & Range & Explanations \\
	\midrule
	LR    & C     & (0.01,10) & Regularization strength \\
	\midrule
	\multirow{7}[2]{*}{XGBoost} & n\_estimators & (3,10) & Number of trees \\
	& max\_depth & (3,10) & Max depth \\
	& learning\_rate & (0.01,0.3) & Learning rate \\
	& subsample & (0.6,1) & Sampling ratio \\
	& colsample\_bytree & (0.6,1) & Sampling feature ratio \\
	& reg\_alpha & (0,1) & $L_1$ regularization  \\
	& reg\_lambda & (0,1) & $L_2$ regularization  \\
	\midrule
	\multirow{4}[2]{*}{TARNet} & hidden\_layer & \{50, 100, 200\} & Dim of hidden layer \\
	& outcome\_layer & \{100, 200\} & Dim of outcome layer \\
	& learning\_rate & \{$1e^{-2}, 1e^{-3}$\} & Learning rate \\
	& batch\_size & \{200, 500\} & Batch size \\
	\midrule
	\multirow{6}[2]{*}{Dragonnet} & alpha & \{0.1, 0.5, 1, 2\} & Weight of treatment loss \\
	& beta  & \{0.1, 0.5, 1, 2\}  & Targeted regularization  \\
	& hidden\_layer &  \{100, 200\} & Dim of hidden layer \\
	& outcome\_layer &  \{100, 200\} & Dim of outcome layer \\
	& learning\_rate & \{$1e^{-2}, 1e^{-3}$\} & Learning rate \\
	& batch\_size & \{200,500\} & Batch size \\
	\bottomrule
\end{tabular}%
}
\end{table}%

\begin{table*}[t]
	\centering
	\caption{Comparison of model performance (measured by PEHE$_{30\%}$ averaged over 10 runs) under various settings (Table omits the corresponding values of $\theta_1 \in \{0.8, 0.95, 1.1\}$ in Setting B). Lower values indicate better performance. }
	\resizebox{2.1\columnwidth}{!}{
	\begin{tabular}{ccccccccccccc}
		\toprule
		\multicolumn{1}{c}{\multirow{2}[4]{*}{\textbf{Model/Setting}}} & \multicolumn{3}{c}{\textbf{Setting A (varying $\xi$)}} & \multicolumn{3}{c}{\textbf{Setting B (varying $\theta_0$)}} & \multicolumn{3}{c}{\textbf{Setting C (varying $\omega$)}} & \multicolumn{3}{c}{\textbf{Setting D (varying $m$)}} \\
		\cmidrule(lr){2-4} \cmidrule(lr){5-7} \cmidrule(lr){8-10} \cmidrule(lr){11-13}          & \textbf{0.8} & \textbf{1.6} & \textbf{2.4} & \textbf{0.4} & \textbf{0.5} & \textbf{0.6} & \textbf{1.2} & \textbf{2.4} & \textbf{3.6} & \textbf{0.1} & \textbf{0.3} & \textbf{0.5} \\
		\midrule
		S-learner & 2.6860  & 2.6162  & 2.6905  & 2.6694  & 3.1238  & 3.6429  & 2.6694  & 2.6394  & 2.6551  & 2.6694  & \textbf{2.8040 } & \textbf{2.9545 } \\
		T-learner & 2.7296  & 2.7995  & 2.8449  & 2.7604  & 3.2095  & 3.6946  & 2.7604  & 2.7251  & 2.7434  & 2.7604  & 2.9870  & 3.0597  \\
		X-learner & 2.7712  & 2.8247  & 2.8632  & 2.7492  & 3.2435  & 3.7478  & 2.7492  & 2.7512  & 2.7441  & 2.7492  & 2.9827  & 3.0822  \\
		R-learner & \textbf{2.6032 } & \textbf{2.5219 } & \textbf{2.4124 } & \textbf{2.6417 } & \textbf{3.1016 } & \textbf{3.5165 } & \textbf{2.6417 } & \textbf{2.6130 } & \textbf{2.6032 } & \textbf{2.6417 } & \textbf{2.8712 } & 3.0388  \\
		U-learner & \textcolor[rgb]{ .051,  .051,  .051}{\textbf{2.5828 }} & \textbf{2.4473 } & \textbf{2.3895 } & \textbf{2.6549 } & \textbf{3.0766 } & \textbf{3.5295 } & \textbf{2.6549 } & \textbf{2.6231 } & \textbf{2.6266 } & \textbf{2.6549 } & 2.8866  & 3.0466  \\
		DR-learner & 2.7772  & 2.8404  & 2.9050  & 2.7483  & 3.2638  & 3.7512  & 2.7483  & 2.7993  & 2.7399  & 2.7483  & 2.9837  & 3.0677  \\
		RA-learner & 2.6700  & \textbf{2.0681 } & \textbf{1.7253 } & 3.0100  & 3.2469  & \textbf{3.4909 } & 3.0100  & 3.0775  & 2.9233  & 3.0100  & 3.0624  & \textbf{2.8635 } \\
		TARNet & \textcolor[rgb]{ .051,  .051,  .051}{\textbf{2.6382 }} & 2.6680  & 2.7507  & \textbf{2.6365 } & \textbf{3.1192 } & 3.5958  & \textbf{2.6365 } & \textbf{2.6301 } & \textbf{2.6090 } & \textbf{2.6365 } & \textbf{2.8707 } & \textbf{3.0030 } \\
		Dragonnet & 2.7930  & 3.1418  & 2.9525  & 2.8236  & 3.3065  & 3.7491  & 2.8236  & 2.8953  & 2.7709  & 2.8236  & 2.9415  & 3.1043  \\
		\bottomrule
	\end{tabular}%
}
	\label{tab:PEHE_table}%
\end{table*}%

\begin{table*}[t]
	\centering
	\caption{Comparison of model performance (measured by ATE$_{30\%}$ averaged over 10 runs) under various settings (Table omits the corresponding values of $\theta_1 \in \{0.8, 0.95, 1.1\}$ in Setting B). Higher values indicate better performance. }
		\resizebox{2.1\columnwidth}{!}{
	\begin{tabular}{ccccccccccccc}
		\toprule
		\multicolumn{1}{c}{\multirow{2}[4]{*}{\textbf{Model/Setting}}} & \multicolumn{3}{c}{\textbf{Setting A (varying $\xi$)}} & \multicolumn{3}{c}{\textbf{Setting B (varying $\theta_0$)}} & \multicolumn{3}{c}{\textbf{Setting C (varying $\omega$)}} & \multicolumn{3}{c}{\textbf{Setting D (varying $m$)}} \\
		\cmidrule(lr){2-4} \cmidrule(lr){5-7} \cmidrule(lr){8-10} \cmidrule(lr){11-13}          & \textbf{0.8} & \textbf{1.6} & \textbf{2.4} & \textbf{0.4} & \textbf{0.5} & \textbf{0.6} & \textbf{1.2} & \textbf{2.4} & \textbf{3.6} & \textbf{0.1} & \textbf{0.3} & \textbf{0.5} \\
		\midrule
		S-learner & 2.7054  & 2.6744  & 2.6295  & 2.8102  & 2.8640  & 3.0842  & 2.8102  & 2.8134  & 2.8096  & 2.8102  & 2.7056  & 2.1225  \\
		T-learner & 2.8102  & 2.7429  & \textbf{2.6856 } & \textbf{2.8841 } & 3.0120  & 3.1529  & \textbf{2.8841 } & 2.8746  & \textbf{2.8818 } & \textbf{2.8841 } & \textbf{2.7606 } & 2.1834  \\
		X-learner & \textbf{2.8181 } & \textbf{2.7536 } & 2.6836  & 2.8765  & \textbf{3.0297 } & \textbf{3.1697 } & 2.8765  & 2.8774  & 2.8716  & 2.8765  & \textbf{2.7596 } & \textbf{2.1867 } \\
		R-learner & 2.7944  & 2.6808  & 2.6557  & 2.8499  & 2.9866  & 3.1343  & 2.8499  & 2.8541  & 2.8550  & 2.8499  & 2.7581  & \textbf{2.1871 } \\
		U-learner & 2.7968  & 2.6824  & 2.6765  & 2.8315  & 2.9834  & 3.1332  & 2.8315  & 2.8409  & 2.8480  & 2.8315  & 2.7567  & 2.1832  \\
		DR-learner & 2.8179  & 2.7358  & 2.6735  & 2.8709  & 3.0226  & 3.1568  & 2.8709  & \textbf{2.8908 } & 2.8697  & 2.8709  & 2.7583  & \textbf{2.1838 } \\
		RA-learner & 1.6375  & 1.5869  & 1.5844  & 1.5675  & 1.6645  & 1.7246  & 1.5675  & 1.5971  & 1.5700  & 1.5675  & 1.5854  & 1.2348  \\
		TARNet & \textbf{2.8421 } & \textbf{2.7537 } & \textbf{2.7042 } & \textbf{2.8783 } & \textbf{3.0296 } & \textbf{3.1775 } & \textbf{2.8783 } & \textbf{2.8778 } & \textbf{2.8813 } & \textbf{2.8783 } & 2.7547  & 2.1757  \\
		Dragonnet & \textbf{2.8477 } & \textbf{2.7622 } & \textbf{2.7143 } & \textbf{2.9005 } & \textbf{3.0469 } & \textbf{3.2046 } & \textbf{2.9005 } & \textbf{2.8865 } & \textbf{2.8991 } & \textbf{2.9005 } & \textbf{2.7722 } & 2.1679  \\
		\bottomrule
	\end{tabular}%
}
	\label{tab:ATE_table}%
\end{table*}%

\begin{figure*}[]
    \centering
      \includegraphics[width=\textwidth]{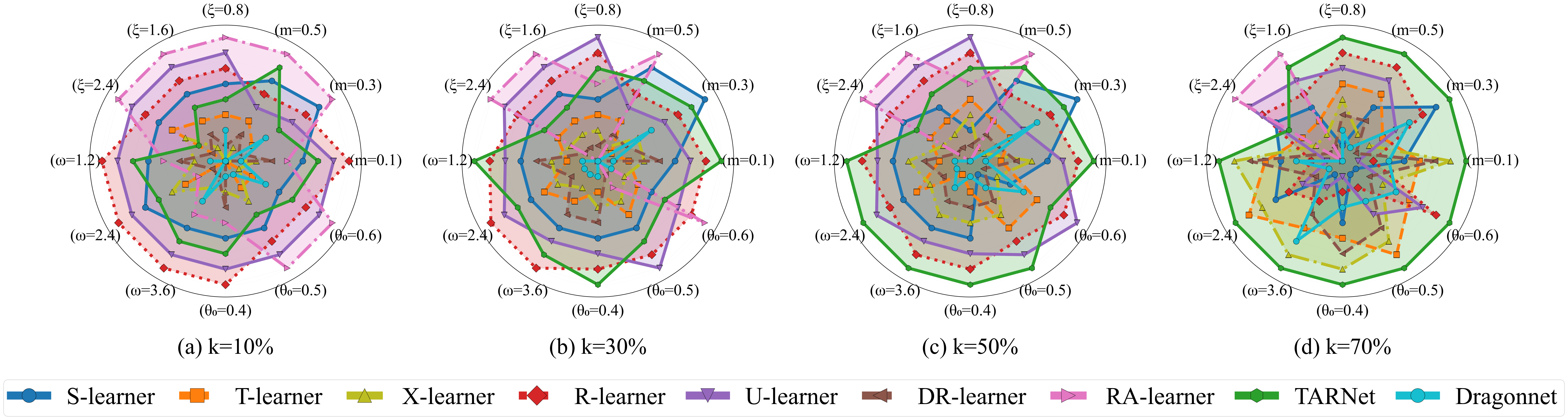}
      \caption{Radar chart of model performance ranked by PEHE$_k$ across different bias settings. Outer points indicate higher ranks.}
      \label{PEHE}
\end{figure*}

\begin{figure*}[]
    \centering
      \includegraphics[width=\textwidth]{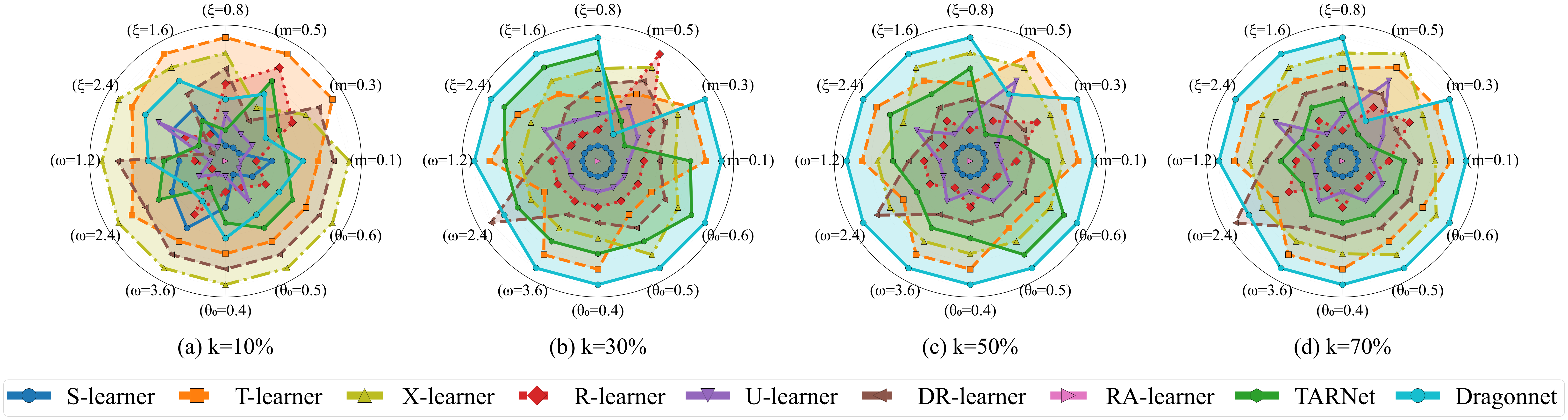}
     \caption{Radar chart of model performance ranked by ATE$_k$ across different bias settings. Outer points indicate higher ranks.}
     \label{ATE}
\end{figure*}

\subsection{Model Performance Under Structural Biases}\label{conclusion_pehe}
\subsubsection{Trade-off between PEHE and ATE in Model Performance} A comparison between \Cref{tab:PEHE_table} and \Cref{tab:ATE_table} reveals a clear decoupling of individual-level precision and population-level accuracy. Specifically, while meta-learners such as the R-learner and U-learner achieve superior PEHE rankings, they frequently suffer from substantial ATE bias. Dragonnet exhibits the opposite trend, providing the most reliable ATE estimates despite its lower PEHE performance. This divergence, consistent across all models except for TARNet (a detailed analysis is provided in \ref{tarnet}), underscores the structural tension between \textit{Uplift Prediction} and \textit{Uplift Targeting} \cite{zhurethinking}. Such results indicate that prioritizing the fit of local fluctuations can inadvertently degrade the global stability of causal estimation.

\subsubsection{Linking Model Design to Performance Gaps.} This performance difference can be attributed to the distinct optimization objectives of these two model classes. 

Explicit effect modeling via objective decomposition is exemplified by meta-learners such as the R-learner, which relies on pseudo-outcomes. In practice, the R-learner is formulated by optimizing a weighted objective function derived from Robinson’s transformation: $$\hat{\tau} = \arg\min_{\tau} \sum_{i=1}^{n} \left[ (Y_i - \hat{m}(X_i)) - \tau(X_i)(T_i - \hat{e}(X_i)) \right]^2$$
where $Y_i$ and $T_i$ are the observed outcome and treatment assignment, respectively. The nuisance components $\hat{m}(X_i) = \mathbb{E}[Y \mid X_i]$ and $\hat{e}(X_i) = P(T = 1 \mid X_i)$ denote the conditional mean outcome and the propensity score. This approach explicitly models the treatment effect $\tau(X_i)$ as a function of the features $X$, enabling the model to focus on the heterogeneity of causal effects across individual characteristics. Furthermore, it incorporates Neyman orthogonality to achieve debiasing. 

Implicit distributional balancing via representation learning, exemplified by Dragonnet, shifts the focus from simple outcome curve-fitting to learning a causally informed latent embedding $\phi(X)$. Building on the dual-head architecture seen in TARNet, Dragonnet integrates propensity score prediction into a joint objective: $$\mathcal{L}_{total} = \frac{1}{n} \sum_{i=1}^n \left( Q(\phi(X_i), T_i) - Y_i \right)^2 + \alpha \mathcal{L}_{CE}$$
where $\mathcal{L}_{CE} = \text{CrossEntropy}(g(\phi(X_i)), T_i)$ is the cross-entropy loss for treatment assignment, and $Q(\cdot)$ and $g(\cdot)$ represent the outcome and propensity heads, respectively. These heads operate on the shared representation layers denoted by $\phi(\cdot)$. The hyperparameter $\alpha$ regulates the trade-off between the two tasks. This multi-task design forces the shared layer $\phi(X)$ to retain information relevant to the treatment assignment mechanism, capturing critical data that a pure regression model might otherwise discard as noise. By coupling the outcome and treatment tasks, the model ensures the latent representation remains grounded in the confounding structure. Ultimately, this approach achieves implicit balancing without the need for explicit reweighting, prioritizing ATE estimation robustness over the fine-grained fitting of individual heterogeneity.

\subsubsection{Why TARNet is Special?}\label{tarnet}
TARNet adopts a shared representation architecture inspired by the T-learner framework, employing two independent regression heads to model treatment-specific potential outcomes. This architecture is formulated by defining the predicted outcome for unit $i$ as: $$\hat{Y}_i = T_i h_1(\Phi(X_i)) + (1 - T_i) h_0(\Phi(X_i))$$
where $T_i \in \{0, 1\}$ denotes the binary treatment assignment, and $h_1(\cdot)$, $h_0(\cdot)$ represent the hypothesis heads corresponding to the treated and control groups.

To optimize the model, the loss function is defined as the mean squared error: $$\mathcal{L} = \frac{1}{n} \sum_{i=1}^{n} (\hat{Y}_i - Y_i)^2$$ This hybrid architecture bridges explicit branching with implicit representation, effectively merging the strengths of both approaches. By pairing deep neural networks’ feature extraction with the structural separation of potential outcome modeling, the model balances PEHE and ATE estimation. This ensures the framework remains globally stable while capturing heterogeneous treatment effects.

\subsubsection{Evaluation with Varied Targeting Fractions.}
To ensure these performance gaps are robust, we extended our evaluation to additional targeting fractions ($k \in \{10\%, 50\%, 70\%\}$). While the polygon shapes in the radar charts (\Cref{PEHE} and \Cref{ATE}) shift slightly with $k$, the relative model rankings remain remarkably consistent. Across all subplots, neural-based models like Dragonnet consistently align better with population-level ATE, whereas meta-learners (e.g., R-learner and U-learner) retain their edge in individual-level PEHE. This stability suggests that our findings reflect inherent model behaviors rather than artifacts of a specific data fraction. We therefore focus our detailed analysis on $k = 30\%$, as it serves as a representative case for how these different designs handle structural biases.

\begin{figure*}[]
    \centering
      \includegraphics[width=\textwidth]{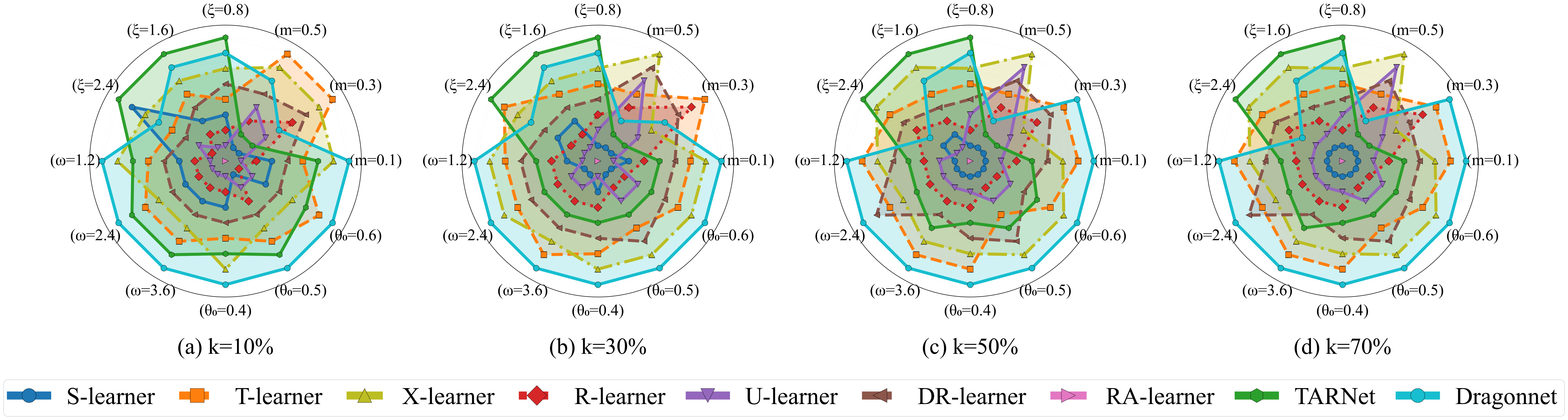}
     \caption{Radar chart of model performance ranked by AUUC$_k$ across different bias settings. Outer points indicate higher ranks.}
     \label{AUUC_radar}
\end{figure*}

\begin{figure*}[]
    \centering
      \includegraphics[width=\textwidth]{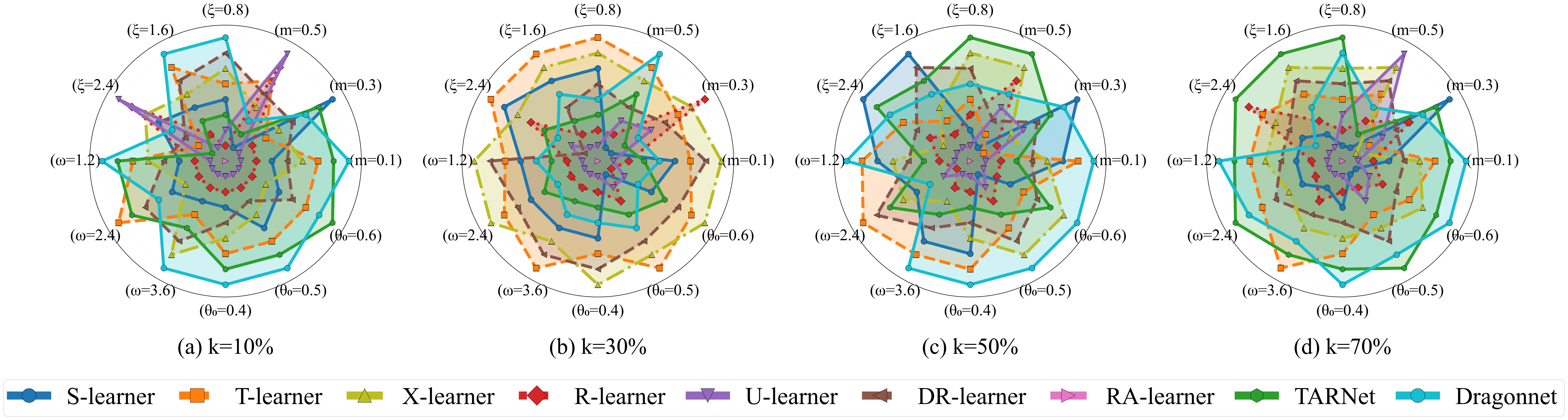}
     \caption{Radar chart of model performance ranked by Qini$_k$ across different bias settings. Outer points indicate higher ranks.}
     \label{Qini_radar}
\end{figure*}

\begin{figure*}[]
    \centering
      \includegraphics[width=\textwidth]{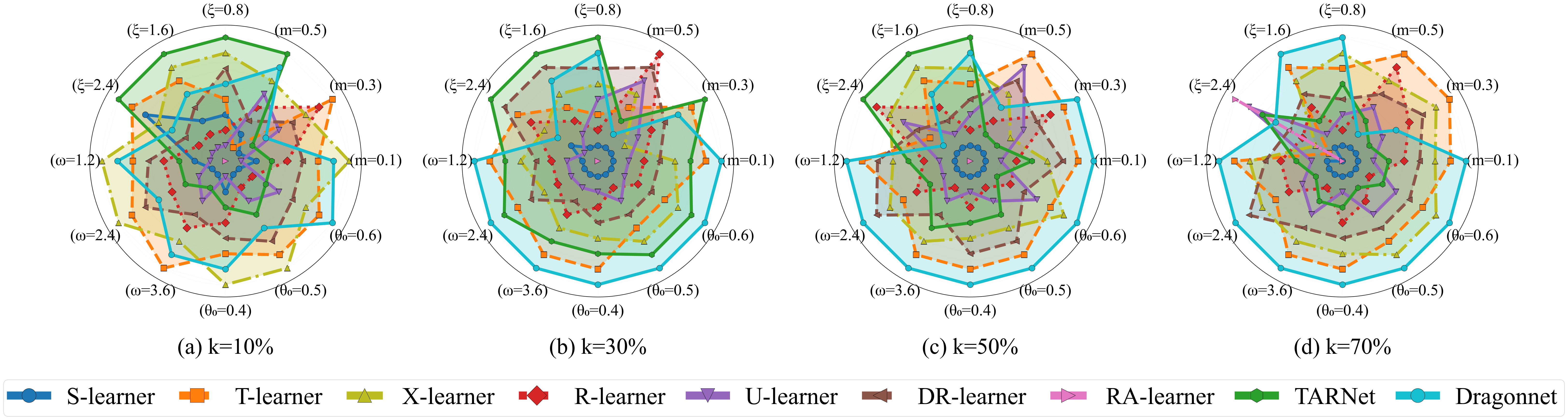}
     \caption{Radar chart of model performance ranked by Uplift$_k$ across different bias settings. Outer points indicate higher ranks.}
     \label{Uplift_radar}
\end{figure*}

\begin{table*}[]
	\centering
	\caption{Comparison of rank correlation for different evaluation metrics (measured by Spearman rank correlation coefficient averaged over 10 runs) under various settings (Table omits the corresponding values of $\theta_1 \in \{0.8, 0.95, 1.1\}$ in Setting B). Higher values indicate better performance. }
	\begin{tabular}{ccccccccccccc}
		\toprule
		\multicolumn{1}{c}{\multirow{2}[4]{*}{\textbf{Metric/Setting}}} & \multicolumn{3}{c}{\textbf{Setting A (varying $\xi$)}} & \multicolumn{3}{c}{\textbf{Setting B (varying $\theta_0$)}} & \multicolumn{3}{c}{\textbf{Setting C (varying $\omega$)}} & \multicolumn{3}{c}{\textbf{Setting D (varying $m$)}} \\
		\cmidrule(lr){2-4} \cmidrule(lr){5-7} \cmidrule(lr){8-10} \cmidrule(lr){11-13}          & \textbf{0.8} & \textbf{1.6} & \textbf{2.4} & \textbf{0.4} & \textbf{0.5} & \textbf{0.6} & \textbf{1.2} & \textbf{2.4} & \textbf{3.6} & \textbf{0.1} & \textbf{0.3} & \textbf{0.5} \\
    \midrule
Uplift$_{10\%}$ & 0.417 & 0.733 & 0.383 & \textbf{0.883} & \textbf{0.967} & \textbf{0.667} & \textbf{0.883} & \textbf{0.783} & 0.550 & \textbf{0.883} & 0.867 & 0.333 \\
AUUC$_{10\%}$ & 0.333 & 0.683 & 0.367 & 0.800 & 0.667 & 0.650 & 0.800 & 0.617 & 0.367 & 0.800 & \textbf{0.933} & \textbf{0.583} \\
Qini$_{10\%}$ & \textbf{0.550} & \textbf{0.917} & \textbf{0.467} & 0.700 & 0.400 & 0.567 & 0.700 & 0.700 & \textbf{0.600} & 0.700 & 0.083 & 0.467 \\
\midrule
Uplift$_{30\%}$ & \textbf{0.950} & 0.850 & 0.483 & \textbf{1.000} & \textbf{0.950} & \textbf{0.983} & \textbf{1.000} & \textbf{0.867} & \textbf{0.983} & \textbf{1.000} & 0.533 & \textbf{0.917} \\
AUUC$_{30\%}$ & \textbf{0.950} & \textbf{0.917} & \textbf{0.783} & 0.900 & 0.933 & 0.867 & 0.900 & 0.817 & 0.950 & 0.900 & \textbf{0.800} & 0.800 \\
Qini$_{30\%}$ & 0.350 & 0.467 & 0.233 & 0.483 & 0.750 & 0.533 & 0.483 & 0.500 & 0.550 & 0.483 & 0.633 & 0.283 \\
\midrule
Uplift$_{50\%}$ & \textbf{0.950} & 0.833 & 0.300 & 0.950 & 0.850 & 0.733 & 0.950 & \textbf{0.983} & \textbf{1.000} & 0.950 & 0.867 & \textbf{0.917} \\
AUUC$_{50\%}$ & \textbf{0.950} & \textbf{0.867} & \textbf{0.567} & \textbf{0.983} & \textbf{0.917} & \textbf{0.950} & \textbf{0.983} & \textbf{0.983} & 0.983 & \textbf{0.983} & \textbf{0.917} & 0.833 \\
Qini$_{50\%}$ & 0.800 & 0.267 & 0.350 & 0.733 & 0.867 & 0.900 & 0.733 & 0.567 & 0.667 & 0.733 & 0.017 & 0.283 \\
\midrule
Uplift$_{70\%}$ & \textbf{0.983} & \textbf{0.967} & -0.167 & 0.983 & 0.950 & 0.983 & 0.983 & \textbf{0.967} & \textbf{0.983} & 0.983 & 0.817 & 0.833 \\
AUUC$_{70\%}$ & 0.833 & 0.817 & \textbf{0.450} & \textbf{1.000} & \textbf{0.983} & \textbf{1.000} & \textbf{1.000} & 0.950 & 0.967 & \textbf{1.000} & \textbf{0.950} & \textbf{0.950} \\
Qini$_{70\%}$ & 0.817 & 0.700 & 0.100 & 0.850 & 0.783 & 0.883 & 0.850 & 0.600 & 0.817 & 0.850 & -0.067 & 0.900 \\
\bottomrule
	\end{tabular}%
	\label{tab:rank}%
\end{table*}%

\subsection{Evaluation Metric Robustness Under Biases}
\subsubsection{Analysis of Evaluation Metrics under Different Biases.} We begin by assessing the structural reliability of the metrics, focusing on the consistency of model rankings across different radar charts (Figures \ref{AUUC_radar}, \ref{Qini_radar} and \ref{Uplift_radar}). The Uplift and AUUC charts exhibit expansive, overlapping boundaries across all bias types, reflecting stable performance rankings. In contrast, the Qini radar charts (\Cref{Qini_radar}) struggle under Selection Bias (Setting A) and Unobserved Confounding (Setting D), where they visibly shrink and become irregular. This divergence suggests that the choice of metric materially affects the assessment of model reliability; it changes how we judge model reliability, with Qini coefficient exhibiting substantially lower stability in biased environments.

\subsubsection{Further Analysis of Metrics Based on Spearman Rank Correlation.}
To evaluate the robustness of standard uplift evaluation metrics, we analyze the Spearman rank correlation of Uplift, AUUC, and Qini with the oracle ATE across four controlled bias settings. This analysis investigates whether these widely adopted metrics consistently recover the oracle model rankings. As illustrated in \Cref{tab:rank}, metrics derived from average treatment effects, such as Uplift and AUUC, demonstrate reliable robustness across various experimental settings. This phenomenon is not limited to a single data split ratio because the Spearman rank correlation coefficients for Uplift/AUUC remain high as $k$ increases. This yields model rankings that closely align with those obtained using ATE, while the Qini coefficient consistently underperforms with lower correlation levels than Uplift/AUUC in most cases.

\subsubsection{Connecting Metric Logic to Robustness Divergence} \Cref{tab:rank} reveals that Uplift and AUUC exhibit strong robustness across different configurations. As detailed in \Cref{Appendix B}, these metrics utilize a mean-based logic by normalizing outcomes ($S_{N_k}^{T}/N_{N_k}^{T}$ and $S_{N_k}^{C}/N_{N_k}^{C}$) to estimate the expected incremental performance per unit. This structural alignment with the ATE definition ($\mathbb{E}[Y \mid T = 1] - \mathbb{E}[Y \mid T = 0]$) may partly explain the observed relative stability. Conversely, the Qini coefficient operates on population-scale logic. $V_i^{QC}$ employs the ratio $\frac{N_i^T}{N_i^C}$ to scale control responses onto the treatment scale. While this balances group sizes, it also introduces significant sensitivity to group imbalance and structural bias. For example, when the control group is small, the scaling factor can amplify estimation error, causing the metric to deviate from the true incremental effect. In summary, while Qini is effective for estimating total business gain, Uplift and AUUC appear more reliable for ranking models under structural bias, as their mean-based logic is less sensitive to the group imbalances we observed.

\section{Conclusion}
In this study, we conducted a comprehensive empirical analysis to evaluate the robustness of uplift models and evaluation metrics under four prevalent marketing biases: selection bias, spillover effects, measurement error, and hidden confounding. Our empirical results reveal several key insights: (1) uplift targeting and uplift prediction are distinct, implying that individual-level precision does not automatically translate into population-level efficacy; (2) the observed robustness of TARNet suggests that its shared representation and head separation may provide a more balanced trade-off, though these findings still require further investigation for more comprehensive validation; (3) observations from our experiments suggest that Uplift and AUUC, which are mathematically analogous to ATE, exhibit relative stability under the bias settings considered in this study. This observation provides a preliminary framework for developing more robust assessment criteria for similar contexts. However, further research is required in settings where the ATE itself may be difficult to identify.

\textit{Limitations and Future Work.} While our work offers practical empirical guidance for uplift modeling in precision marketing, it has several limitations that suggest avenues for future exploration. Firstly, we focus on four common biases, but other important challenges in personalized marketing, such as data imbalance \citep{nyberg2021uplift}, limited supervision \citep{panagopoulos2024uplift}, and carryover effects \citep{shi2023dynamic}, remain unexamined. Investigating how these factors influence model performance and the robustness of evaluation metrics would be a valuable next step. Secondly, while this study involves a rigorous empirical analysis, the scope is primarily limited to the Hillstrom benchmark dataset. This choice was intended to provide a controlled experimental environment with reduced noise. However, the findings may not fully generalize to the substantial variability found in diverse real-world industrial contexts. Therefore, future research is needed to evaluate the applicability of this framework across a broader range of industrial datasets. Thirdly, this study evaluates nine representative uplift models rather than more recent architectural developments. This choice is based on the premise that many newer models are iterative modifications of these established paradigms. By examining these foundational models, we seek to understand how standard architectural frameworks are affected by specific biases. This analysis may serve as a useful reference for researchers when considering different architectural paradigms. Fourthly, spillover effects in this study are primarily predicated on Euclidean distance rather than more intricate network structures. This design is a deliberate attempt to isolate the potential impact of spillover effects on model performance, thereby reducing the likelihood of interference from other confounding factors. We acknowledge that this approach is a simplification. Therefore, future research could explore more sophisticated spillover mechanisms that better reflect real-world complexities while continuing to refine methods for decoupling these effects from extraneous variables. Fifthly, although our findings demonstrate that Uplift and AUUC exhibit high evaluation stability, a multi-metric approach is recommended in practical applications to ensure a comprehensive performance profile. Furthermore, the continuous development of novel metrics remains essential to further enhance assessment reliability across diverse and complex bias scenarios.

\begin{acks}
    This work was supported by the National Natural Science Foundation of China (No. 62302310), the Startup Funds of Great Bay University (No. YJKY250111), and the Innovative Team Program for Regular Universities in Guangdong Province (No. 2025KCXTD031).
\end{acks}



\bibliographystyle{ACM-Reference-Format}
\balance

\bibliography{KDD26/reference}


\appendix

\section{Details of Evaluation Metrics}\label{Appendix B}


We evaluate models using both oracle metrics (relying on counterfactual information) and practical metrics (computed from observed data). Let $k$ denote the percentile of the total $N$ ranked units (e.g., $k = 30\%$), and $N_k = N \cdot k$ denote the number of top-$k$ ranked units.

\textit{Oracle metrics.}  
The metrics PEHE$_k$ and ATE$_k$ serve as gold standards for uplift model evaluation but are unavailable in real-world applications because they require counterfactual outcomes.
\textbf{PEHE$_k$} measures the root mean squared error between predicted and true individual-level uplift values among the top-$k$ customers (ranked by the model $\hat{\tau}$) \cite{hill2011bayesian,shalit2017estimating,zhurethinking}:
\begin{equation}
	\mathrm{PEHE}_k = \sqrt{\frac{1}{N_k} \sum_{i=1}^{N_k} \left( \hat{\tau}(x_i) - \tau(x_i) \right)^2}.
\end{equation}
This metric reflects how well a model $\hat{\tau}$ identifies individuals with the highest true uplift values, making it an oracle indicator for the \emph{uplift prediction} task.

\textbf{ATE$_k$} measures the true population-level uplift values among the top-$k$ customers (ranked by the model $\hat{\tau}$):
\begin{equation}
	\mathrm{ATE}_k = \frac{1}{N_k} \sum_{i=1}^{N_k} \tau(x_i).
\end{equation}
It represents the potential average profit that could be achieved by targeting the top-$k$ customers, serving as an oracle metric for the \emph{uplift targeting} task.

\textit{Practical metrics.}
In practice, we cannot observe counterfactuals; thus, we can only rely on surrogate metrics computed from observed data. Let $N_i^T$ and $N_i^C$ be the numbers of treated and control units among the top-$\frac{i}{N}$ ranked customers, $S_i^T$ and $S_i^C$ be their cumulative observed outcomes, and $\mathbf{1}_{\{\cdot\}}$ be the indicator function; then we define the following necessary terms for the $i$-th individual:
\begin{equation*}
	\begin{aligned}
		&N_{i}^T = \sum_{j = 1}^{i} \mathbf{1}_{\{T_j = 1\}}, \quad
		&&N_{i}^C = \sum_{j = 1}^{i}  \mathbf{1}_{\{T_j = 0\}}, \\ 
		&S_{i}^T = \sum_{j = 1}^{i}  \mathbf{1}_{\{T_j = 1\}} y_j, \quad
		&&S_{i}^C = \sum_{j = 1}^{i}  \mathbf{1}_{\{T_j = 0\}} y_j.
	\end{aligned}
\end{equation*}

\textbf{Uplift$_k$} directly estimates the difference in average cumulative observed outcomes between treatment and control groups within the top-$k$ ranked customers:
\begin{equation}
	\mathrm{Uplift}_k = \frac{S_{N_k}^T}{N_{N_k}^T} - \frac{S_{N_k}^C}{N_{N_k}^C}.
\end{equation}

\textbf{AUUC$_k$} (Area Under the Uplift Curve) integrates the uplift curve over the top-$k$ ranked customers \citep{rzepakowski2012decision,zhurethinking}:
\begin{equation}
	\mathrm{AUUC}_k = \sum_{i=1}^{N_k-1} \frac{V^{\mathrm{UC}}_{i} + V^{\mathrm{UC}}_{i+1}}{2},
\end{equation}

where the uplift curve value function at the $i$-th individual is defined as: 
$$V^{\mathrm{UC}}_{i} = \left( \frac{S_{i}^T}{N_{i}^T}
-
\frac{S_{i}^C}{N_{i}^C} \right) (N_{i}^T + N_{i}^C)$$
\textbf{Qini$_k$} quantifies the improvement in targeting performance over a random strategy using the Qini curve \citep{surry2011quality,radcliffe2007using,diemert2018large,devriendt2018literature,belbahri2021qini}:
\begin{equation}
	\mathrm{Qini}_k = {\mathrm{Area}(S_{\mathrm{model}}, k) - \mathrm{Area}(S_{\mathrm{random}}, k)},
\end{equation}
where $S_{\mathrm{model}}$ and $S_{\mathrm{random}}$ denote the ranking induced by $\hat{\tau}$ and by random selection, respectively. For a given ranking strategy $S$, the area is computed as:
\begin{equation}
	\mathrm{Area}(S, k) = \sum_{i=1}^{N_{k}-1} \frac{V^{\mathrm{QC}}_{i}(S) + V^{\mathrm{QC}}_{i+1}(S)}{2},
\end{equation}
where the Qini curve value function at the $i$-th individual is defined as:
$$V^{\mathrm{QC}}_{i}(S) = S_{i}^T - \frac{N_{i}^T}{N_{i}^C}S_{i}^C$$.

\section{Results and Analysis of the Bank Dataset}
\label{Appendix E}

Complementing the Hillstrom dataset in the main text, this section reports the PEHE and ATE results on the Bank dataset \cite{bank_marketing_222}. Results are shown in \Cref{bank_pehe} and \Cref{bank_ate}. These supplementary empirical results are largely consistent with the model-level observations in the main content. In particular, TARNet achieves consistently strong performance on both PEHE and ATE, suggesting that its shared representation with treatment-specific heads provides a favorable trade-off between individual-level uplift prediction and population-level targeting. The Bank results also exhibit a PEHE--ATE discrepancy, although this pattern is less pronounced than that observed on the Hillstrom dataset. Specifically, the model rankings induced by PEHE and ATE are not always identical, indicating that accurate individual-level uplift estimation does not necessarily translate into optimal population-level targeting performance. Meanwhile, TARNet remains a stable exception, further supporting our observation that structural separation of potential outcome modeling can improve robustness under biased data-generating conditions.

\begin{table*}[htbp]
  \centering
  \caption{Results on the Bank dataset: Comparison of model performance (measured by PEHE$_{30\%}$ averaged over 10 runs) under various settings (Table omits the corresponding values of $\theta_1 \in \{0.8, 0.95, 1.1\}$ in Setting B). Lower values indicate better performance.}
  \label{bank_pehe}
    \resizebox{2.1\columnwidth}{!}{
    \begin{tabular}{ccccccccccccc}
    \toprule
    \multicolumn{1}{c}{\multirow{2}[4]{*}{\textbf{Model/Setting}}} & \multicolumn{3}{c}{\textbf{Setting A (varying $\xi$)}} & \multicolumn{3}{c}{\textbf{Setting B (varying $\theta_{0}$)}} & \multicolumn{3}{c}{\textbf{Setting C (varying $\omega$)}} & \multicolumn{3}{c}{\textbf{Setting D (varying $m$)}} \\
\cmidrule{2-13}          & \textbf{0.8} & \textbf{1.6} & \textbf{2.4} & \textbf{0.4} & \textbf{0.5} & \textbf{0.6} & \textbf{1.2} & \textbf{2.4} & \textbf{3.6} & \textbf{0.1} & \textbf{0.3} & \textbf{0.5} \\
    \midrule
    S-learner & 2.2471  & 2.2272  & 2.1129  & 2.2202  & 2.2815  & 2.2588  & 2.2202  & 2.2202  & 2.2202  & 2.2202  & 2.4187  & 3.7783  \\
    T-learner & \textbf{1.7829 } & 1.8215  & \textbf{1.8108 } & \textbf{1.8589 } & \textbf{1.8772 } & 1.8964  & \textbf{1.8589 } & \textbf{1.8589 } & \textbf{1.8589 } & \textbf{1.8589 } & \textbf{2.1192 } & \textbf{3.4795 } \\
    X-learner & 1.8170  & 1.7917  & 1.8111  & \textbf{1.8467 } & 1.8943  & \textbf{1.8845 } & \textbf{1.8467 } & \textbf{1.8468 } & \textbf{1.8471 } & \textbf{1.8467 } & 2.1314  & \textbf{3.4734 } \\
    R-learner & 2.6312  & 2.4902  & 2.8622  & 2.7786  & 2.5291  & 2.6807  & 2.7786  & 2.7297  & 2.7483  & 2.7786  & 2.6756  & 3.5028  \\
    U-learner & 2.9636  & 73.8624  & 9.2050  & 2.8027  & 3.0159  & 2.8829  & 2.8027  & 2.7270  & 2.7808  & 2.8027  & 2.6614  & 3.5086  \\
    DR-learner & \textbf{1.7623 } & \textbf{1.7885 } & \textbf{1.7962 } & 1.8715  & \textbf{1.8911 } & 1.9145  & 1.8715  & 1.8853  & 1.8698  & 1.8715  & \textbf{2.1123 } & 3.4928  \\
    RA-learner & 7.6948  & 8.5461  & 9.0887  & 6.3567  & 6.4315  & 6.5907  & 6.3567  & 6.3567  & 6.3567  & 6.3567  & 6.4451  & 4.5412  \\
    TARNet & \textbf{1.6358 } & \textbf{1.6961 } & \textbf{1.6759 } & \textbf{1.7038 } & \textbf{1.6666 } & \textbf{1.7574 } & \textbf{1.7038 } & \textbf{1.6833 } & \textbf{1.6946 } & \textbf{1.7038 } & \textbf{2.0124 } & \textbf{3.4895 } \\
    Dragonnet & 1.9384  & \textbf{1.7656 } & 2.1815  & 1.8614  & 2.0536  & \textbf{1.7834 } & 1.8614  & 2.0070  & 1.8739  & 1.8614  & 2.4142  & 3.7479  \\
    \bottomrule
    \end{tabular}}%
  \label{tab:addlabel}%
\end{table*}%

\begin{table*}[htbp]
  \centering
  \caption{Results on the Bank dataset: Comparison of model performance (measured by ATE$_{30\%}$ averaged over 10 runs) under various settings (Table omits the corresponding values of $\theta_1 \in \{0.8, 0.95, 1.1\}$ in Setting B). Higher values indicate better performance.  }
  \label{bank_ate}
    \resizebox{2.1\columnwidth}{!}{
    \begin{tabular}{ccccccccccccc}
    \toprule
    \multicolumn{1}{c}{\multirow{2}[4]{*}{\textbf{Model/Setting}}} & \multicolumn{3}{c}{\textbf{Setting A (varying $\xi$)}} & \multicolumn{3}{c}{\textbf{Setting B (varying $\theta_{0}$)}} & \multicolumn{3}{c}{\textbf{Setting C (varying $\omega$)}} & \multicolumn{3}{c}{\textbf{Setting D (varying $m$)}} \\
\cmidrule{2-13}          & \textbf{0.8} & \textbf{1.6} & \textbf{2.4} & \textbf{0.4} & \textbf{0.5} & \textbf{0.6} & \textbf{1.2} & \textbf{2.4} & \textbf{3.6} & \textbf{0.1} & \textbf{0.3} & \textbf{0.5} \\
    \midrule
    S-learner & 3.0847  & 3.1177  & 3.1857  & 3.1153  & 3.1205  & 3.1582  & 3.1153  & 3.1153  & 3.1153  & 3.1153  & 2.9675  & 2.3754  \\
    T-learner & 3.2149  & 3.2156  & 3.2218  & 3.2097  & 3.2299  & 3.2420  & 3.2097  & 3.2097  & 3.2097  & 3.2097  & 3.0827  & \textbf{2.5112 } \\
    X-learner & 3.2037  & 3.2138  & \textbf{3.2231 } & \textbf{3.2325 } & 3.2420  & \textbf{3.2589 } & \textbf{3.2325 } & \textbf{3.2331 } & \textbf{3.2328 } & \textbf{3.2325 } & \textbf{3.0864 } & 2.5099  \\
    R-learner & 3.1319  & 3.1332  & 3.0265  & 2.8459  & 2.9489  & 2.9341  & 2.8459  & 2.8629  & 2.8611  & 2.8459  & 2.8831  & 2.5098  \\
    U-learner & 3.0855  & 2.8689  & 2.5634  & 2.8195  & 2.9061  & 2.9042  & 2.8195  & 2.8202  & 2.8121  & 2.8195  & 2.8601  & 2.5020  \\
    DR-learner & \textbf{3.2165 } & \textbf{3.2215 } & \textbf{3.2531 } & 3.2267  & \textbf{3.2422 } & 3.2557  & 3.2267  & 3.2242  & 3.2179  & 3.2267  & 3.0701  & \textbf{2.5103 } \\
    RA-learner & 0.9467  & 1.0779  & 1.0881  & 0.2731  & 0.2895  & 0.2442  & 0.2731  & 0.2731  & 0.2731  & 0.2731  & 0.3039  & 0.1294  \\
    TARNet & \textbf{3.3334 } & \textbf{3.3274 } & \textbf{3.3303 } & \textbf{3.3538 } & \textbf{3.3703 } & \textbf{3.3699 } & \textbf{3.3538 } & \textbf{3.3609 } & \textbf{3.3567 } & \textbf{3.3538 } & \textbf{3.1745 } & \textbf{2.5281 } \\
    Dragonnet & \textbf{3.2752 } & \textbf{3.2776 } & 3.1827  & \textbf{3.2779 } & \textbf{3.2966 } & \textbf{3.3038 } & \textbf{3.2779 } & \textbf{3.2830 } & \textbf{3.2753 } & \textbf{3.2779 } & \textbf{3.0841 } & 2.4515  \\
    \bottomrule
    \end{tabular}}%
  \label{tab:addlabel}%
\end{table*}%

\end{document}